\documentclass{article}
\usepackage{spconf,amsmath,graphicx}
\usepackage{hyperref}
\usepackage[table,xcdraw]{xcolor}
\usepackage{multirow}
\usepackage{float}                 
\usepackage{tabularx}
\title{Multimodal Emotion Recognition and Sentiment Analysis in Multi-Party Conversation Contexts}

\name{
    Aref Farhadipour, Hossein Ranjbar, Masoumeh Chapariniya,
    Teodora Vukovic, Sarah Ebling, Volker Dellwo }
\address{
    Department of Computational Linguistics, University of Zurich, Zurich, Switzerland }
\begin{document}
%
\maketitle
\begin{abstract}
Emotion recognition and sentiment analysis are pivotal tasks in speech and language processing, particularly in real-world scenarios involving multi-party, conversational data. This paper presents a multimodal approach to tackle these challenges on a well-known dataset. 
We propose a system that integrates four key modalities/channels using pre-trained models: RoBERTa for text, Wav2Vec2 for speech, a proposed FacialNet for facial expressions, and a CNN+Transformer architecture trained from scratch for video analysis. Feature embeddings from each modality are concatenated to form a multimodal vector, which is then used to predict emotion and sentiment labels. The multimodal system demonstrates superior performance compared to unimodal approaches, achieving an accuracy of 66.36\% for emotion recognition and 72.15\% for sentiment analysis. 

\end{abstract}

\section{Introduction}
\label{sec:intro}

With the rapid growth of technology, the role of machines in various sectors, such as resource management, entertainment, and human assistance, has become increasingly prominent. In these applications, machines are typically provided with data from the real world, and they are expected to respond appropriately. The input data can be diverse, including text, audio, video, or multimodal information.\\
Each type of input requires specific processing techniques, and each modality has distinct challenges. In multimodal data processing, selecting and implementing an effective method for each modality is critical to achieving optimal performance.

One significant challenge that has gained considerable attention in recent years is enhancing the intelligence of Human-Machine Interaction (HMI). Users interacting with machines often wish to communicate their emotional states, making it imperative for the system to accurately perceive and respond to these emotions to facilitate a high level of intelligent interaction. Therefore, developing emotion and general sentiment recognition systems has become a crucial area of research.

Based on the theory of emotions, the emotion recognition task identifies the inner sense based on the seven basic emotions. However, sentiment analysis deals with labelling the general opinion expressed as positive, negative, or neutral, for example, in customer feedback analysis.

Emotion recognition in HMI involves multiple modalities, such as text, facial expressions, voice, and body movements, representing different aspects of human emotions. These modalities complement each other; for example, voice tone, facial expressions, and gestures can collectively modify or amplify the emotional content of spoken words. 

However, designing multimodal emotion recognition and sentiment analysis systems poses challenges due to the scarcity of large, high-quality multimodal datasets. Most of the current research is based on the IEMOCAP dataset \cite{busso2008iemocap}, which is recorded under controlled studio conditions. While this dataset provides a solid foundation, real-world emotion recognition presents more significant challenges, especially in multi-party conversational settings. For this reason, we employ the MELD dataset \cite{poria2018meld} in this work, as it offers a more realistic setting with varied background conditions, multiple speakers, and environmental noise, providing a more challenging yet practical scenario for emotion recognition. 

In this work, we propose a multimodal system designed for emotion recognition and sentiment analysis. This system is built upon four core components, each addressing a distinct modality. Specifically, we utilize Wav2Vec2 \cite{baevski2020wav2vec} for voice, RoBERTa \cite{liu2019roberta} for text, a proposed FacialNet for facial expressions, and a CNN+Transformer architecture trained from scratch for video analysis. 

The remainder of this paper is organized as follows: Section \ref{sec:whsp} reviews related work in emotion recognition and sentiment analysis. Section \ref{sec:mthd} describes the dataset and the proposed system architecture. Experimental results are presented in Section \ref{sec:exp}, followed by conclusions in Section \ref{sec:conc}.\\

\section{Previous Work}
\label{sec:whsp}

While some approaches focus on unimodal emotion recognition \cite{far2023facial}, recent advancements have favoured the integration of multiple data modalities and machine learning techniques. These multimodal methods aim to address the limitations of unimodal analysis by combining information from diverse sources to enhance emotion recognition performance.

Zheng et al. \cite{zheng2023facial} proposed a two-stage framework focusing on facial sequences. Their approach employs multi-task learning to generate emotion-informed visual representations. Similarly, Hu et al. \cite{hu2023supervised} developed the supervised adversarial contrastive learning framework, which incorporates contextual adversarial training alongside a Dual-LSTM structure to enhance emotion recognition from textual data.

Yun et al. \cite{yun2024telme} introduced the Teacher-leading Multimodal Fusion Network for emotion recognition, which employs cross-modal knowledge distillation. In this method, a language model is used to improve the performance of non-verbal modalities. Fu et al. \cite{fu2024ckerc} presented a bi-model system for emotion that uses prompts to generate commonsense knowledge for interlocutors based on historical utterances. This model incorporates an interlocutor commonsense identification task to fine-tune the system for identifying implicit speaker cues.

Chudasama et al. \cite{chudasama2022m2fnet} introduced a multimodal fusion network, which extracts emotion-relevant features from visual, audio, and textual modalities. Their model uses a multi-head attention-based fusion mechanism and introduces a novel feature extractor trained with adaptive margin-based triplet loss for the audio and visual modalities. 

In another novel approach, Wu et al. \cite{wu2024beyond} developed a system that applies large language models to speech emotion recognition. This method converts speech characteristics into natural language descriptions and incorporates them into text prompts, allowing for multimodal emotion analysis without modifying the model’s architecture.

Multimodal sentiment analysis in conversations has also gained traction as researchers seek to overcome the limitations of unimodal approaches. Poria et al. \cite{poria2018meld} compared various models, including Text-CNN, LSTM, and DialogueRNN, finding that multimodal fusion outperformed unimodal methods, with DialogueRNN excelling in multi-party situations. Shah et al. \cite{shah2023ensemble} took an ensemble learning approach to sentiment analysis, integrating RoBERTa for text and Wav2Vec2 for audio. Their study explored various ensemble techniques to improve sentiment classification accuracy through multimodal fusion.


\section{Methodology}
\label{sec:mthd}

In the proposed system, we used several pre-trained models. The use of pre-trained models is driven by the fact that available multimodal emotion recognition and sentiment analysis datasets are often limited in size, making transfer learning an effective strategy to overcome this data scarcity challenge. Various fusion techniques are commonly employed in multimodal systems, including sensory, feature, and score fusion and based on prior work, feature fusion has consistently demonstrated superior results \cite{farhadipour2024compara}.

\begin{figure}[t]
    \centering
    \includegraphics[width=0.5\textwidth, height=0.4\textheight]{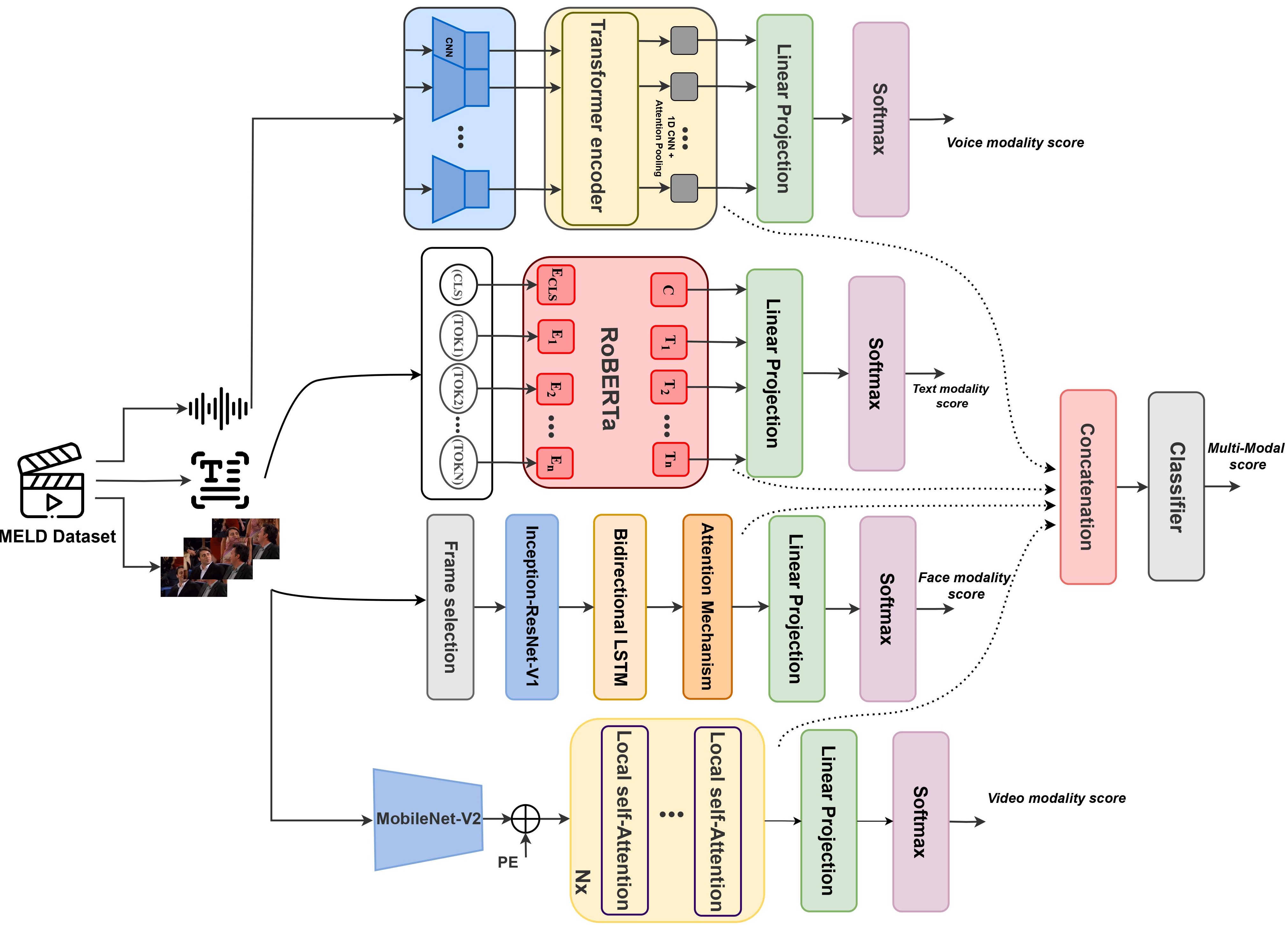}
    \caption{Proposed system with unimodal and multimodal feature fusion strategies.}
    \label{fig:F0}
\end{figure}
Figure \ref{fig:F0} depicts the proposed system's architecture for emotion recognition and sentiment analysis tasks. Two approaches are explored: unimodal and multimodal fusion strategies. In the multimodal mode, the feature vectors extracted from each system are concatenated to make a multimodal vector.

For the voice modality, Wav2Vec2 is applied in an audio classification framework. The model is initially trained on three emotion recognition datasets and later fine-tuned on the MELD dataset. A similar approach is adopted for RoBERTa, pre-trained on 58 million tweets for emotion recognition using the TweetEval benchmark before fine-tuning on MELD \cite{barbieri2020tweeteval}. The combined representation is then used to make final decisions regarding emotion and sentiment via an additional classifier. The multimodal feature vector is classified using a multi-layer perceptron.

The rest of this section outlines the materials and methods used in this study. We introduce the MELD dataset and briefly overview the four models employed for each modality: Wav2Vec2, RoBERTa, FacialNet, and CNN+Transformer.

\subsection{Dataset}
\label{ssec:data}

MELD is a comprehensive multimodal dataset designed to advance emotion recognition and sentiment analysis tasks. It incorporates multiple modalities, including text, audio, and visual features \cite{poria2018meld}. One of the unique aspects of MELD is that it is derived from dialogues in the popular TV series \textit{Friends}, which provides emotionally diverse conversations between multiple speakers in dynamic, real-world-like settings. This makes MELD particularly valuable for real-world emotion recognition tasks, offering over 1,400 dialogue instances and more than 13,000 individual utterances.

Each utterance in the dataset is annotated with seven emotion and three sentiment labels. The seven emotions include anger, fear, disgust, joy, sadness, surprise, and neutrality, while sentiment labels are positive, negative, and neutral.

\subsection{RoBERTa}
\label{ssec:hub}

RoBERTa (Robustly Optimized BERT Pretraining Approach) is a transformer-based language model built upon BERT (Bidirectional Encoder Representations from Transformers). It is designed to enhance performance by optimizing training strategies, including larger batch sizes, more training data, and removing the next sentence prediction objective. RoBERTa \cite{liu2019roberta} employs a masked language model, which learns contextualized word representations by predicting randomly masked tokens within a sentence based on surrounding context. By leveraging its bidirectional context understanding, RoBERTa captures subtle linguistic cues such as tone, word choice, and sentence structure, contributing to detecting underlying emotions.

\subsection{Wav2Vec2}
\label{ssec:Wav2Vec2}

Wav2Vec2 \cite{baevski2020wav2vec} has revolutionized speech processing by applying self-supervised learning directly to raw audio data. The model consists of a convolutional feature extractor and a transformer-based contextual network that models temporal dependencies in the input. During training, segments of the input speech are masked, and the network is tasked with predicting the masked portions, thus learning rich speech representations.

Wav2Vec2 has demonstrated strong performance in various tasks, including speech recognition and emotion classification, due to its ability to model complex acoustic features directly from raw waveform data \cite{baevski2020wav2vec}.

\subsection{CNN+Transformer}
\label{ssec:WavLM}

We developed a model from scratch that extracts visual features not limited to facial expressions to handle video understanding. While previous works primarily focus on facial analysis, our approach captures broader visual cues, including the speaker's posture, gestures, the presence of other individuals, and background elements, which may contribute valuable context to emotion recognition.

Our architecture comprises three key components: a spatial modelling module, a temporal modelling module, and a classifier. We use MobileNetV2 \cite{sandler2018mobilenetv2} as a feature extractor for spatial modelling. This architecture is chosen for its efficiency and low latency, making it well-suited for real-time applications.

For temporal modelling, we employ a transformer-based approach \cite{vaswani2017attention, ranjbar2024continuous, rios2023multimodal} that captures temporal dependencies between frames. Given that long-range dependencies may not always be necessary for emotion recognition, we implement a local attention mechanism \cite{child2019generating} to focus on relevant temporal information. This module consists of two transformer layers, each with eight attention heads and a hidden size 1280. Finally, the extracted features are passed through a neural network classifier with a hidden size 512 to predict emotion labels.

\subsection{FacialNet}
\label{ssec:Whisper}

The facial model in this work employs a frame selection strategy to extract relevant face frames from videos using a two-stage process \cite{zheng2023facial}. First, TalkNet, an active speaker detection model, combines multimodal rules to detect potential speaker faces. Then, InfoMap clustering and face matching against a character face library are used to isolate the actual speaker's face sequence.

For feature extraction, we utilize an InceptionResNetv1 model \cite{szegedy2017inception}, pre-trained on the CASIA-WebFace dataset \cite{yi2014learning} that is initially for the face recognition task. This model is responsible for processing face frames and generating visual representations. A bidirectional LSTM is employed to classify the extracted features. This LSTM applies an attention mechanism to compute a context vector passed through a fully connected layer to produce the final classification output.

\begin{table}[]
\centering
\caption{Results in percentage for emotion recognition in unimodal mode}
\label{tab:t1}
\resizebox{0.40\textwidth}{!}{
\begin{tabular}{c|cccc}
         & Text  & Voice & Face  & Video \\ \hline
Accuracy & 64.34 & 51.49 & 22.61 & 36.14
\end{tabular}
}
\caption{Results in percentage for sentiment analysis in single-modality mode}
\label{tab:t2}
\centering
\resizebox{0.40\textwidth}{!}{
\begin{tabular}{c|cccc}
         & Text  & Voice & Face  & Video \\ \hline
Accuracy & 69.21 & 56.20 & 38.98 & 42.51
\end{tabular}
}
\end{table}

\section{Experimental Results}
\label{sec:exp}

The proposed systems were evaluated on the MELD dataset, focusing on two tasks: emotion recognition and sentiment analysis. In the emotion recognition task, the model predicts one of the seven basic emotions from the input multimedia data, including voice, video frames, facial expressions, and the corresponding text of utterances. The system classifies each input file for sentiment analysis as positive, negative, or neutral.

The models were fine-tuned with a batch size of 16 over five epochs to avoid over-fitting, using  AdamW for optimization \cite{loshchilov2017decoupled}. The video modality, which was trained from scratch, underwent training for 10 iterations. Tables \ref{tab:t1} and \ref{tab:t2} report the results for emotion recognition and sentiment analysis, respectively, for each unimodal system. As expected, the system performed better in sentiment analysis compared to emotion recognition. This can be attributed to the reduced number of sentiment classes, as multiple emotions are grouped into broader sentiment categories.

The image-based systems, including face and video processing, faced significant challenges due to variations in background, camera angles, occlusions, and multiple faces in a frame. Consequently, the accuracy for these modalities was lower. Specifically, the video analysis system, which aimed to capture general visual information and gestures, achieved an accuracy of 36.14\% for emotion recognition and 42.51\% for sentiment analysis. However, Multiple facial frames extracted from each video exhibit diverse emotional expressions yet are assigned a single emotion label. The processing of several frames consequently for a single label leads to low accuracy of FacialNet, in such a way that emotion recognition achieved  22.61\% and 38.98\% for sentiment analysis.

Despite the challenges of multi-party conversations and environmental noise, the voice modality achieved 51.49\% accuracy for emotion recognition and 56.20\% for sentiment analysis. The text modality demonstrated the highest performance, with 64.34\% accuracy for emotion recognition and 69.21\% for sentiment analysis, underscoring the effectiveness of linguistic cues in these tasks.

\begin{table}[]
\centering
\caption{Emotion recognition in feature fusion}
\label{tab:t3}
\resizebox{0.4\textwidth}{!}{
\begin{tabular}{l|c}
Modalities                  & Accuracy (\%) \\ \hline
Voice + Face + Video        & 48.16         \\
Text + Face + Video         & 65.98         \\
Text + Face + Voice         & 66.25         \\
Text + Video + Voice         & 66.29         \\
Text + Face + Video + Voice  & 66.36        
\end{tabular}
}
\centering
\caption{Sentiment analysis in feature fusion}
\label{tab:t4}
\resizebox{0.4\textwidth}{!}{
\begin{tabular}{l|c}
Modalities                  & Accuracy (\%) \\ \hline
Voice + Face + Video        & 49.58         \\
Text + Video + Voice          & 71.76         \\
Text + Video + Face          & 71.80         \\
Text + Face + Voice          & 71.84         \\
Text + Video + Face + Voice & 72.15        
\end{tabular}
}
\end{table}
The primary challenge with the text modality was handling conversational speech by target speakers that make various affect expressions corresponding to each audience. However, this modality is generally cleaner compared to others, allowing the system to learn the parameters for each class effectively. Each modality may excel in different aspects of the dataset, as each one captures a unique facet of emotion. This variability highlights the potential benefits of a multimodal approach, where integrating multiple modalities can address individual weaknesses.

In a multimodal strategy, the objective is to construct a multimodal vector for each input file. The classifier's role is to process this multimodal vector, learn from labeled samples during training, and identify the crucial components of the multimodal representation. Various integration strategies have been explored, with their performance for emotion recognition detailed in Table \ref{tab:t3}.

Table \ref{tab:t3} presents the results of different integration approaches for sentiment analysis, including both 3-modality and 4-modality fusion. The text modality proves to be the most effective, indicating its role as a primary modality. The highest accuracy, 66.36\%, was achieved by integrating all four modalities, which represents a 2.02\% improvement over the best-performing in unimodal mode. This result demonstrates that the classifier effectively leverages the information from multiple modalities.

For sentiment analysis, the performance results of different integration approaches are shown in Table \ref{tab:t4}. The best performance, with an accuracy of 72.15\%, was achieved through the integration of all four modalities. In this evaluation, the multimodal sentiment analysis outperforms the unimodal by 2.94\% for accuracy.

\section{Conclusion}
\label{sec:conc}

Emotion recognition and sentiment analysis are closely related and challenging tasks in pattern recognition. With its multi-party and conversational nature, the MELD dataset presents significant challenges due to the extensive visual and auditory variability in each sample. This variability results in differing performance for each modality across files, highlighting the need for multimodal approaches in such scenarios.

In this work, we utilized pre-trained models including RoBERTa, Wav2Vec2, InceptionResNet, and also developed a model based on a CNN+Transformer architecture trained from scratch to capture comprehensive information for video analysis. By extracting feature embeddings from each modality and concatenating them into a multimodal representation, we achieved superior performance compared to unimodal approaches. Our evaluation demonstrates the effectiveness of multimodal approaches for emotion recognition and sentiment analysis in near-to-real-world situations. As future trends, segmenting the speaker out of frame could improve two visual systems, and a multi-network strategy based on arousal and valence to recognize the effects could be beneficial.

\bibliographystyle{IEEEbib}
\bibliography{Template}

\end{document}